\newcommand{\CLASSINPUTtoptextmargin}{72pt}

\newcommand{\CLASSINPUTbottomtextmargin}{54pt}
\documentclass[conference]{IEEEtran}

\usepackage{subcaption}
\usepackage{hyperref}
\usepackage[usenames, dvipsnames]{color}

\usepackage{cite}

\ifCLASSINFOpdf
   \usepackage[pdftex]{graphicx}
   \usepackage[pdftex]{rotating}
\else
   \usepackage[dvips]{graphicx}
\fi
\usepackage{amsmath}
\usepackage{algorithm}
\usepackage{algpseudocode}

\usepackage{array}
\usepackage{url}

\begin{document}

\title{Procedurally Provisioned Access Control for Robotic Systems} 

\author{
    \IEEEauthorblockN{Ruffin White and Henrik I. Christensen}
        \IEEEauthorblockA{Contextual Robotics Institute\\
        UC San Diego, California, USA} 
    \and
    \IEEEauthorblockN{Gianluca Caiazza and Agostino Cortesi}
        \IEEEauthorblockA{Ca' Foscari University of Venice\\
        Venezia VE, Italy} 
}

\maketitle

\begin{abstract}
Security of robotics systems, as well as of the related middleware infrastructures, is a critical issue for industrial and domestic IoT, and it needs to be continuously assessed throughout the whole development lifecycle. The next generation open source robotic software stack, ROS2, is now targeting support for Secure DDS, providing the community with valuable tools for secure real world robotic deployments. In this work, we introduce a framework for procedural provisioning access control policies for robotic software, as well as for verifying the compliance of generated transport artifacts and decision point implementations.

\end{abstract}

\begin{IEEEkeywords}
Cryptobotics, Cybersecurity, Networked Robots, Industrial Robots, Robot Safety, Middleware, ROS2, Secure DDS 
\end{IEEEkeywords}

\section{Introduction}
\label{par:introduction}

Industry 4.0 represents a shift in the future towards ubiquitous connected robotic systems. To cope with the requirement for custom device-tailored solutions, a significant number of different IoT platforms has emerged in the wild \cite{Mineraud2016}. We can easily observe how these are more oriented to the so called Consumer Internet of Things (IoT) rather than to Industrial IoT (IIoT) solutions. However, the high growth rate of consumer solutions in addition to the steady development of Industrial applications has rapidly overcome the security measures that were deployed originally for \emph{one-purpose} devices.

\begin{figure}
    \centering
        \includegraphics[page=2,
            width=\linewidth,
            trim= 0 70 225 0,
            clip]
            {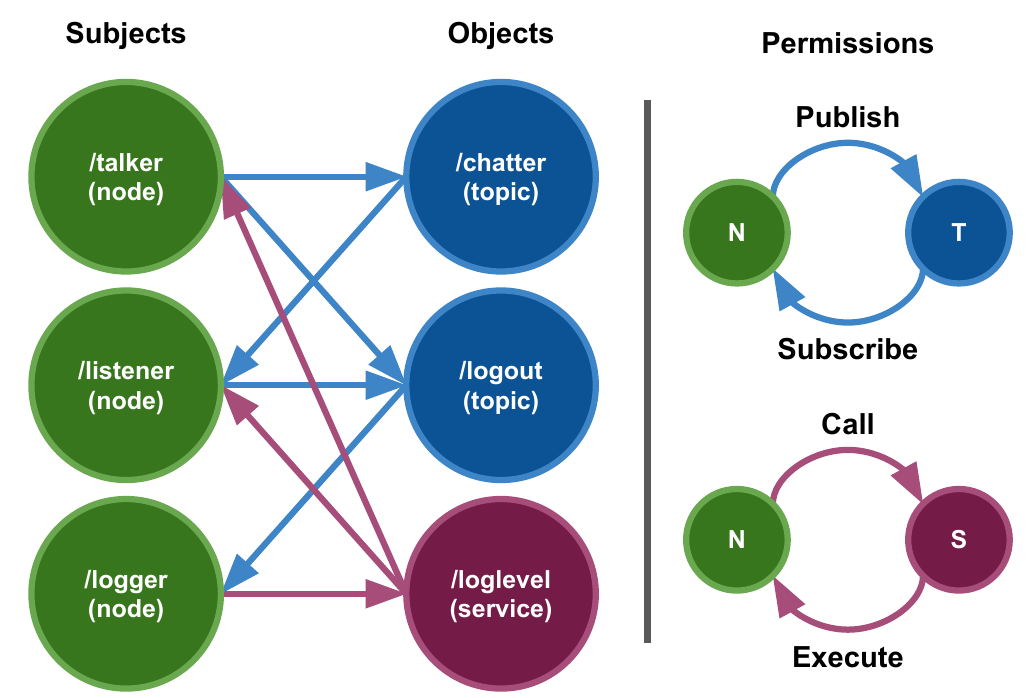}
    \caption{A semantic representation of an example computational graph, often implemented via robotic middleware, modeled as a set of bipartite graphs for each sub-system. Shared vertices among the bigraphs are encompassed by a set of participating nodes. For access control, subject permissions are visualized by directional edges, reflecting the duality of the asymmetric exchange between roles, as opposed to the flow of information.}
    \label{fig:bipartite_graph}
\end{figure} 

As discussed by Morante \emph{et al.} \cite{Morante2015}, the emerging security problems that the field of robotics faces now are similar to those presented for computers in the early days of the Internet. In fact, robotic networks are generally subject to classical cyber-physical attacks such as: Denial of Service (DoS), eavesdropping, tampering, repetition, spoofing, etc.  It's tempting, but we can't simply apply the same counter measures as are used in general computers. In fact, although robotic networks and computer networks may seem similar, they are greatly different \cite{Cardenas}. In robotic networks we need to consider different tradeoffs between safety critical corner cases, and security and real-time constraints that are not present in other systems.

In robotic networks, we can find a significant number of devices that need to operate in conjunction with each other. In such cases, as discussed by Dieber \emph{et al.} \cite{Dieber2017}, the information security pillars of Confidentiality, Integrity and Availability (CIA) that are generally applied to secure information systems assume different priorities.

Generally, in robotic middleware, we consider a subset of the CIA pillars' scope:  confidentiality corresponds to privacy, authenticity coincides with integrity, while availability is necessary for network protocols and hardware redundancy.

To provide those properties in middleware systems such as the Robotic Operating System (ROS) \cite{quigley2009ros} or Data Distribution Service (DDS) \cite{pardo2003}, each node in the distributed computation graph or participant on the data bus is attributed to an identity. This is commonly done by provisioning each node with a X.509 certificate, signed by a trusted Certificate Authority (CA), using an established Public Key Infrastructure (PKI).

Furthermore, depending on the implementation, in addition to the identification mechanism, access control is also deployed. As we will see in Section \ref{par:background},  access control may also be enforced by allocating given roles or attributes to participants in control policy.

However, the additional tasks and commitment that are imposed upon developers to properly generate, maintain and distribute the number of signed public certificates, ciphered private keys, and access control documents attributed to every identity within the distributed network can prove beyond tedious and be error prone. In fact, the additional complexity and scalability of these networks makes the secure orchestration of those systems a demanding process. 

The main contribution of this paper is: in order to mitigate the risks of improper provisioning, we contribute a set of tools to provide users with an automated approach for systematic generation and verification of necessary cryptographic artifacts in a familiar, yet extendable, meta-build system layout via workspaces and plugins.

\subsection*{Overview}
\begin{itemize}
    \item \ref{par:background} \textbf{Related Work:} Discussion of robotic network vulnerability, as well as the limits of the existing approaches on security research in robotic frameworks.
    \item \ref{par:approach} \textbf{Approach:} Outlines general structure of the presented framework, design mechanisms and development choices.
    \item \ref{par:results} \textbf{Results:} Evaluates the capability of the proposed framework with regards to integrity verification of securing a general robotic application.
    \item \ref{par:conclusion} \textbf{Conclusion and Future Work:} Includes a discussion of the presented work and possible improvements to future robotic security by means of more powerful policy definition mechanisms. 
\end{itemize}

\begin{figure}
    \centering
    \begin{subfigure}[b]{0.5\linewidth}
        \includegraphics[page=4,
            width=\textwidth,
            trim= 100 0 100 0,
            clip]
            {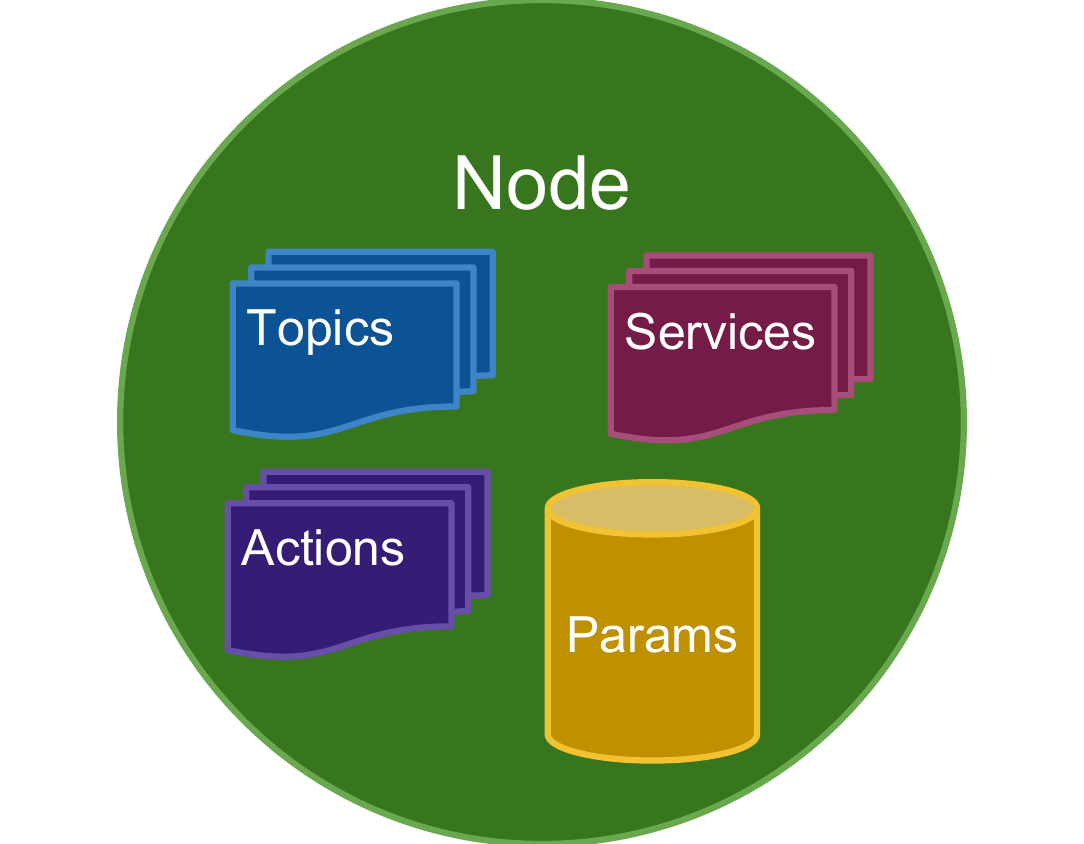}
    \end{subfigure}%
    \begin{subfigure}[b]{0.5\linewidth}
        \includegraphics[page=5,
            width=\textwidth,
            trim= 100 0 100 0,
            clip]
            {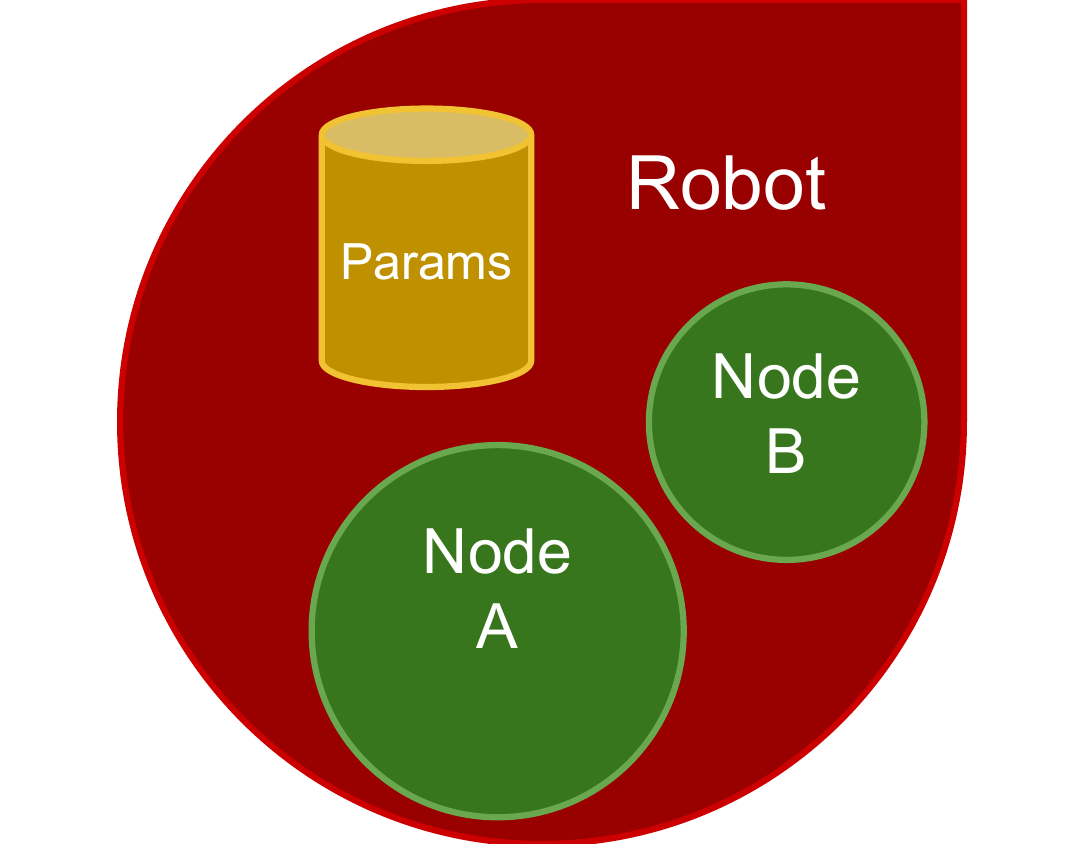}
    \end{subfigure}
    \caption{High level overview of the subsystem for ROS middleware: Nodes organized under a global namespace hierarchy may possess a set of subsystem types such as topics, parameters, services or actions. These subsystems are addressable by other nodes via namespaces that can be orthogonal to the originating node's, but in practice often coincide for organizational clarity. Global parameters are also made available, while in practice simply hosted via another dedicated node.}
    \label{fig:subsystems}
\end{figure}

\section{Related Work}
\label{par:background}

In the early days, robotic systems were simply intended as `\emph{physically}' enhanced computers without specific constraints or limitations. Since the goal was to develop the fastest, lightest and most practical solution for prototyping and deploying products, overall, often the security component and the associated privacy risks were overlooked.

As in the early era of the Internet for Personal Computers (PC), the possibility of a robotic system being compromised was neglected since physical access to the robot was required. For example, in ROS1 the subsystem components communication and parameters handling were designed without considering any security or privacy measures at all (see Figure \ref{fig:subsystems}).  However, now the large scale introduction of robotic applications into daily life coupled with the further adoption of wireless remote access mechanisms has made the possibility of compromised applications a widespread concern.

In this regard, Denning \emph{et al.} \cite{Denning2009} presented an interesting analysis of future risks and general problems deriving from the use of household robots. They analyzed three consumer robots, focusing on their exploitable features. The authors have stated several novel questions to be posed when evaluating future products for both security and privacy design. 

Morante \emph{et al.} \cite{Morante2015} dubbed this new research area which unifies cyber safety and best practices in developing and distributing robot software as \emph{cryptobotics}. In their analysis, by presenting some real-world examples, they pointed out several fields in robotics where security and privacy are critical.

Lee \emph{et al.} \cite{Lee2012} presented security enhancements for the medical field in 'Interoperable Telesurgery Protocol' (ITP) by adding X.509 certificates and encryption by means of TLS in the TCP communication channel. Their novel approach, named Secure ITP, represents one of the first applications of what we discuss in Section \ref{par:introduction} about security measures in robotic frameworks such as the Data Distribution Services (DDS).

With a similar goal, Huang \emph{et al.} \cite{huang2014} have developed \texttt{ROSRV} a runtime verification framework for robotic applications on top of ROS1. ROSRV addresses the security of the ROS computational graph by putting itself in between nodes and master. In this way, if necessary it can amend potentially 'blacklisted' malicious messages in addition to enforcing a manually defined IP based access control policy. There are some drawbacks to this approach: since their access control mechanism relies on IP addresses it remains vulnerable to attacks coming from processes that are running on the same IP. Also, the defined policy profiles are not scalable to different network topologies that may need to be rewritten. 

D\'oczi \emph{et al.} \cite{Doczi2016} proposed for a medical surgery robot based on ROS1 the introduction of authentication and authorization (AA) mechanisms that rely on an additional AA-node to verify the identity of a node in the graph. Their goal was to overcome the delay limitation of other ROS1 oriented security solutions in mission-critical devices. Still, their solution introduced a strong Single Point of Failure (SPOF) - the AA-node - and demanded non trivial setup of the static AA infrastructure.  

As reported by Portugal \emph{et al.} \cite{portugal2017} in recent years, several initiatives have been created to support security in ROS. SROS1 has been proposed as addition to the ROS1 API and ecosystem to support cryptography and security \cite{white2016sros} \cite{white16roscon}; SRI's Secure ROS provided an alternative version of core ROS packages to enable secure communication by means of IPSec, still its IP was based on ROSRV and it suffered the same drawbacks discussed above \cite{SecureROS}; Secure-ROS-Transport sought to enhance the ROS application-level architecture by adding cryptography to the communication channel and an authentication server similar to D\'oczi \emph{et al.} \cite{dieber2016application,breiling2017secure}. 

In addition, we conducted an in deep analysis on how decoupling the provisioning model and improving the policy syntax in ROS1, as well as introducing the ideas presented in this paper, can benefit the robotic community \cite{Koubaa2018}. In fact, all the solutions that have been developed so far require a certain degree of knowledge of the whole network topology.

Interestingly, McClean \emph{et al.} \cite{McClean2013} have shown how exploits and malfunctions debugging in robotic frameworks remains a particularly challenging task. To the same end, Cortesi \emph{et al.} \cite{CortesiFC13} discussed the use of semantics-based static analysis techniques for software verification. Overall, considering how challenging and error-prone this process remains, the use of automatic analysis techniques might help more readily disclose unexpected software behaviours. The use of pluggable security strategies derived from other generated pub/sub frameworks such as Data Distribution Services (DDS) Security might also follow suit, as might Specification\footnote{\href{http://www.omg.org/spec/DDS-SECURITY}{DDS Security: omg.org/spec/DDS-SECURITY}} from the Object Management Group (OMG), as now used in SROS2.

\section{Approach}
\label{par:approach}

As stated in Section \ref{par:introduction}, we seek to mitigate the risks imposed from improper provisioning of robotic middleware credentials that could otherwise compromise system security. To achieve this, we procedulize the provisioning process of all transport artifacts via build automation. Such compilation is made possible by defining an intermediate representation to express the higher level semantics of general permission policies, enabling the compiler to abstract away lower level cryptographic operations. This approach also affords administrators to design policy profiles that are agnostic to the deployed transport, facilitating further consistency of security permissions across transport type, version and vendor.

Our approach contributes two original complementary tools to be used to describe and automate the process for secure and access-controlled communication in data-driven middleware.  The first tool consists of a syntactic language to succinctly describe policy profiles for subjects including any rules for objects while establishing their respective first and second order priorities. The second tool builds off the first, and consists of a cryptographic tool chain for compiling a global symbolic policy representation resulting in individual transport artifacts as required to deploy each subject. 

\subsection{ComArmor}
\begin{figure}
    \centering
        \includegraphics[page=9,
            width=\linewidth,
            trim= 0 120 300 0,
            clip]
            {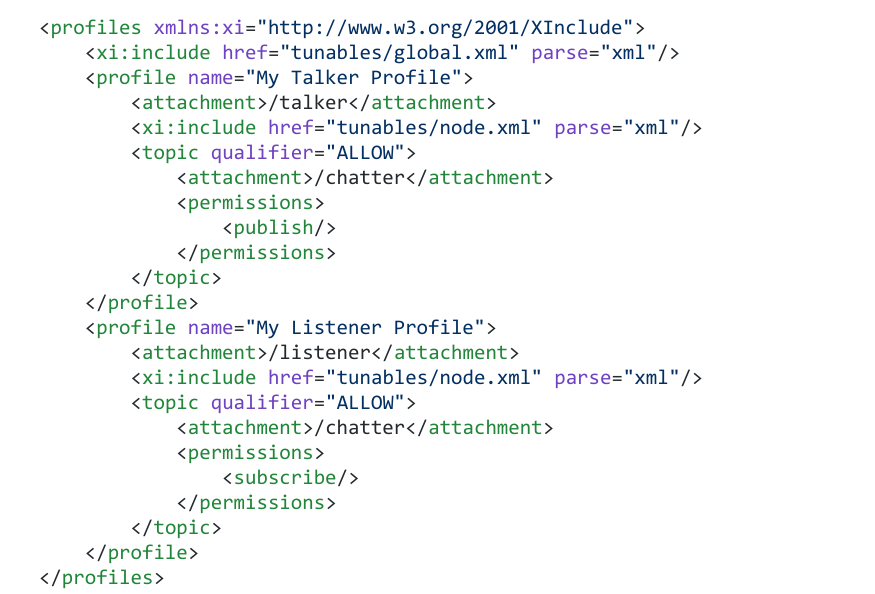}
    \caption{A minimal access control policy for the talker listener example formulated in ComArmor's profile language. Profiles define the scope of rules to bind to a subject via attachment expressions. Rules also make use of attachments to fixate onto objects that are applicable. Profiles may be nested to afford recursive inclusions, affording a high level of composability and reuse of common sub-policies across multiple subjects.}
    \label{fig:comarmor_talker_listener}
\end{figure}

ComArmor\footnote{ComArmor Project: \href{https://github.com/comarmor/comarmor}{github.com/comarmor/comarmor}} is a profile configuration language for defining Mandatory Access Control (MAC) policies for communication graphs. ComArmor is akin to other MAC systems, but rather than defining policy profiles for Linux security modules as with AppArmor\footnote{AppArmor Project: \href{https://gitlab.com/apparmor/apparmor}{gitlab.com/apparmor/apparmor}}\cite{bauer2006paranoid}, ComArmor defines policy profiles for armoring communications, as the project name's alliteration plays upon. ComArmor provides a formalized XML based markup for specifying governance, and accompanied XML Schema Definition (XSD) for validation. These policies, constructed from hierarchical nesting of compositional profiles that bind objects to subjects with prescribed permissions via  attachment expressions, are later read by meta-build stages to procedurally generate end use transport credentials. An excerpt from an example ComArmor profile is shown in Figure \ref{fig:comarmor_talker_listener}.

Borrowing design patterns from the AppArmor community, ComArmor provides an equivalent profile concept, but as opposed to attaching to a process by its executable's path, ComArmor attaches profiles to subjects by their Uniform Resource Identifier (URI), e.g. a node namespace in ROS. A profile encapsulates an un-ordered set of rules, child profiles and includes statements recursively importing more of the former. A nested grand child profile is only made applicable if all of its parent profiles are applicable as well.

Defined rules either allow or deny a specified set of permissions for a given object by URI attachment, e.g. a topic namespace in ROS. ComArmor also works under the same MAC assumptions in AppArmor, i.e. deny by default, where access to any resource or action necessitates it first be explicitly allowed in the policy. Additionally, given deny supersedes any allow, an applicable allow rule alone is insufficient as the absence of any precedent deny rule must also be satisfied. In this way, users can curtail policies with blanketed allow rules over subspaces of an object, but then punch holes in those subspaces, thus revoking specific access to unique resources.

Compared to other more general formats, such as eXtensible Access Control Markup Language (XACML), ComArmor takes an approach that is more straight forward in horizontally transferring permission polices onto the computation graph, where objects are essentially channels on a data buss. Additionally, ComArmor is meant to be compactly human readable while also remaining easily machine generatable. However, we would still like to eventually supply a translation compiler to piggyback on the additional static analysis tools available for XACML, affording more formal verification methods.

\subsection{Keymint}

\begin{figure*}[ht]
    \centering
        \includegraphics[page=10,
            width=\linewidth,
            trim= 0 0 0 55,
            clip]
            {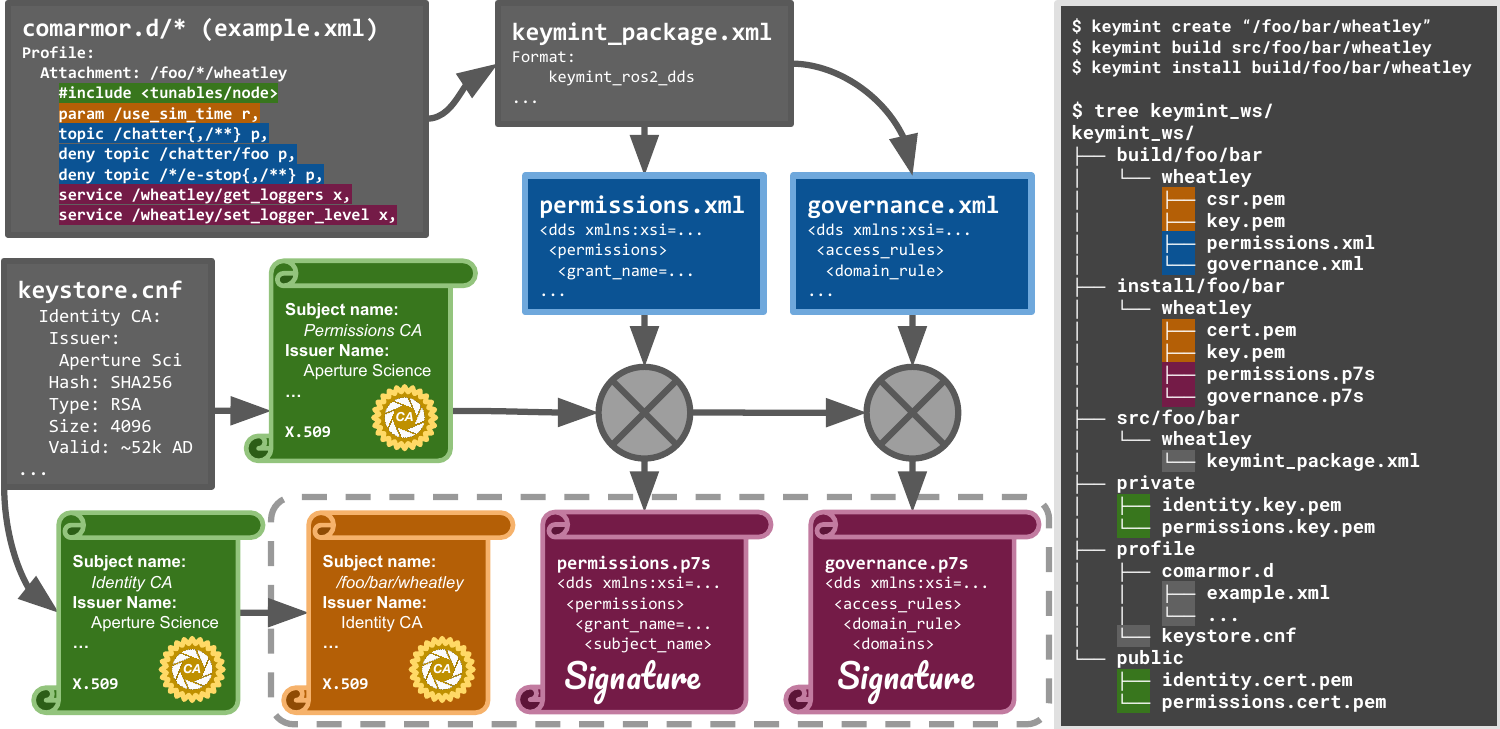}
    \caption{Flow chart visualization of keymint keystore pipeline. Global environment for the keymint workspace is set via keystore config, defining originating hierarchy of trust; i.e. Certificate Authorities. Upon package creation, workspace profiles are used to provision the policy for the target subject; e.g. filtering applicable permissions via the ComArmor plugin. The resulting artifacts for secure \& controlled transport are staged within separate build and install phases, enabling users to customize final documents via entrypoint plugins etc, resembling community's familiar workflow for building and installing source packages.}
    \label{fig:keymint_pipline}
\end{figure*}

Keymint\footnote{Keymint Project: \href{https://github.com/keymint/keymint\_tools}{github.com/keymint/keymint\_tools}} is a framework for generating cryptographic artifacts used in securing middleware systems like ROS, DDS, etc. Keymint is akin to other meta-build systems, but rather than compiling source code and installing executables in workspaces as with Ament\footnote{Ament Project: \href{https://github.com/ament/ament\_tools}{github.com/ament/ament\_tools}}, Keymint mints keys and notarizes documents in keystores, as the project name's alliteration again plays upon. Keymint provides users pluggable tools for automating the provision process for customizing PKI artifacts used with SROS, or Secure DDS plugins.

Keymint's approach in minting cryptographic artifacts resembles that of other common meta-build systems, like Ament, used to compile binary artifacts from source code. Similarly, users create keymint\_packages within an workspace initialized by a keymint\_profile; a package being a structured source manifest describing how and what artifacts are to be generated for an identity, while the workspace provides a tunable profile to adjust the global build context for all packages. In addition, Keymint shares a staggered development cycle, where a workspace is $initialized$, $built$, and $installed$. While each stage in the cycle is subject to the behavior of the plugin invoked as determined by the package's declared format. An example workflow with Keymint is shown in Figure \ref{fig:keymint_pipline}.

While the Keymint library and CLI are intended to be both transport and policy format agnostic, and instead simply operate upon source packages in a workspace containing public and private resources, plugins for ComArmor and ROS2/DDS are included by default. Future policy acquisition plugins for XACML and mySQL may also be added for more advanced policy management. The ComArmor profile and ROS2/DDS build plugins work together with Keymint's Keymake compiler to gather the applicable policy from the package's URI and compile it into an intermediate representation before installing the generated PKI and fixating the permission and governance files via SMIME.

\begin{algorithm}
    \caption{DDS Security v1.0 Default Access Control Logic}
    \label{fig:dds_security_plugin}
    \begin{algorithmic}[1] 
        \Procedure{Evaluate}{$permissions, subject$}
            \For{$grant$ \textbf{in} $permissions$}
                \State $match \gets grant.subject\_name.match(subject)$
                \State $valid \gets grant.validity(current\_date\_time)$
                \If{$match$ \textbf{and} $valid$}
                    \State $qualifier \gets$ \Call{CheckRules}{$rules$, $subject$}
                    \If{$qualifier$ \textbf{is} $None$}
                        \State \textbf{return} $grant.default$
                    \Else
                        \State \textbf{return} $qualifier$
                    \EndIf
                \EndIf
            \EndFor
            \State \textbf{return} $ERROR$
        \EndProcedure
        \Function{CheckRules}{$rules$, $subject$}
            \For{$rule$ \textbf{in} $rules$}
                \State $domain \gets subject.domain$ \textbf{in} $rule.domainSet$
                \State $criteria \gets rule.get(subject.action.type)$
                \State \Comment{Action types: $publish, subscribe, relay$}
                \State $match \gets$ \Call{CheckCriteria}{$criteria$, $subject$}
                \If{$domain$ \textbf{and} $match$}
                    \State \textbf{return} $rules.qualifier$
                    \State \Comment{Qualifier types: $ALLOW, DENY$}
                \EndIf
            \EndFor
            \State \textbf{return} $None$
        \EndFunction
        \Function{CheckCriteria}{$criteria$, $subject$}
            \For{$criterion$, $i$ \textbf{in} $criteria.criterions$}
                \State $matches[i] \gets$ \textbf{any} $(criterion.match(subject))$
                \State \Comment{Criterion types: $topics, partitions, tags$}
            \EndFor
            \State \textbf{return} \textbf{all} $(matches)$
        \EndFunction
    \end{algorithmic}
\end{algorithm}


Essentially, this automates many of the delicate steps in correctly formulating the policy as to be compliant for the transport specific format, the evaluation circuit for DDS Security which is described in Algorithm \ref{fig:dds_security_plugin}, not to mention the additional details in X.509 certificate and keypair generation. Given the mantra that security and usability must go hand in hand, Keymint provides a conservative default bootstrapped workspace suitable for basic users, in which the only configuration required on part of the user is to provide an initial ComArmor profile for the targeted deployment. This in itself is a task that can be automated via training as demonstrated and further exemplified in Section \ref{par:results}.

In addition to compliance, the Keymint policy compilation process can also ensure the transport artifacts result in a faithful interpretation of the original symbolic policy. For example, ComArmor's un-ordered rules sets and deny overrides must be considered accordingly when translating to Secure DDS default plugin permission structure given that its Policy Decision Point (PDP) evaluates upon the first found matching rule in an ordered list. Thus ComArmor deny rules must be arranged in the list as to always be considered first for a given object. Additionally, only the minimally applicable subset of the global ComArmor policy is finally embedded into an individual subject's credentials.

These optimizations are perhaps two of many to consider, with additional ones including: policy compression via folding of collapsible rules that share compatible criteria, thus saving payload overhead in secure transport handshaking; perhaps another ordered prioritization of rules, i.e. quickening average handshakes by placing more frequently requested rules further ahead in the list for faster lookup. Either of these could be beneficial for real time communication that must be adapted to support security overhead.

We have delayed implementing such further optimization until ROS2's DDS namespace mapping is declared stable. As of this writing, ROS2 Ardent has proved cumbersome to regulate due to the fact ROS subsystem namespaces are split across both DDS topics and partitions while additionally prefixed with the subsystem type's action. This mapping approach is also shown be susceptible in section \ref{par:results}. At present, controlling for ROS2 subsystems via DDS security criteria are almost all orthogonal in nature given their entangled relation, thus necessitating a larger number of individual rules than would be otherwise expected. E.g. even ROS2 topics must not share the same DDS publish rule given the chance the union of topic namespace prefixes and names would allow crosstalk between namespaces: 

$$[/foo/bar, /baz/spam] \longrightarrow [/baz/bar, /foo/spam]$$

Abstracting policy definitions away from such complex entanglements is perhaps yet another reason for relying on using intermediate representations and compilers to preform the task on behalf of the system administrator. With ROS2 Bouncy migrating to a more one to one mapping between DDS topics and ROS2 subsystems, compilation will become more straightforward, leaving DDS partitions available as higher level criteria for expressing resource policy definitions as described in section \ref{par:future_work}.

With ComArmor and Keymint enabling repeatable and reproducible cryptographic artifacts, revision controlling the source configurations now becomes both rational and elegant. With this, as with the AppArmor community and Debian packaging, we envision further adoptions to afford ROS package maintainers the  opportunity to provide default configurations, audited and maintained by the community and domain experts.

Anticipating further development of static analysis or manifests of the topology of a system employing orchestration tools and upstarts, using Keymint it is possible to pre-provision all necessary artifacts for deployment.  Alternatively, Keymint's API could be called dynamically to generate artifacts on the fly, as required when remapping subsystem namespaces using the ROS2 launch orchestration.

Public key infrastructure as we know it has remained relatively unchanged for many years, at least as it is used in industry.  With rapid advances in cryptographic research, new and more powerful cryptographic mechanisms such as Ciphertext-Policy Attribute-Based Key Encapsulation Mechanisms (CP-AB-KEM)\cite{hohenberger2014online} or other functional encryption schemes that could afford roboticists more flexible and secure access control definitions.  With Keymint, we have abstracted the policy from the transport-specific details, so with the ratification and industrial adoption of new paradigms, it is possible to simply upgrade the Keymint Keymake compiler to support newer artifact types.


\begin{sidewaysfigure*}[p]
    \centering
    \begin{subfigure}[b]{\linewidth}
        \includegraphics[page=1,
            width=\textwidth,
            trim= 35 15 35 35,
            clip]
            {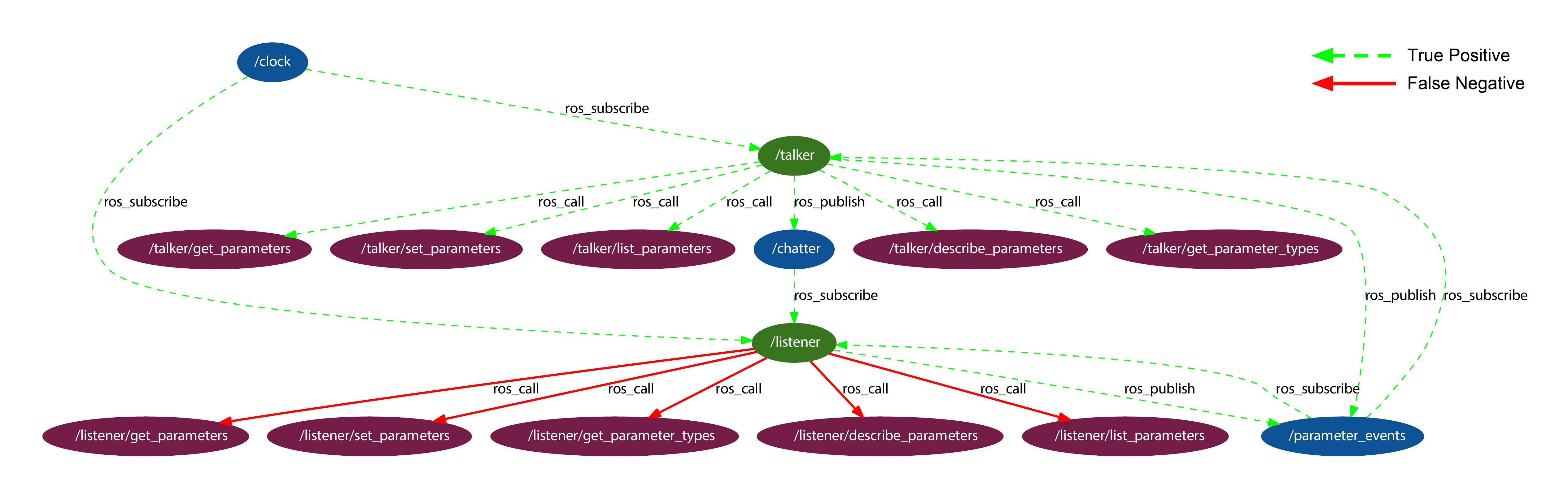}
    \end{subfigure}
    \rule[0.5ex]{\linewidth}{1pt}
    \begin{subfigure}[b]{\linewidth}
        \includegraphics[page=1,
            width=\textwidth,
            trim=35 15 35 15,
            clip]
            {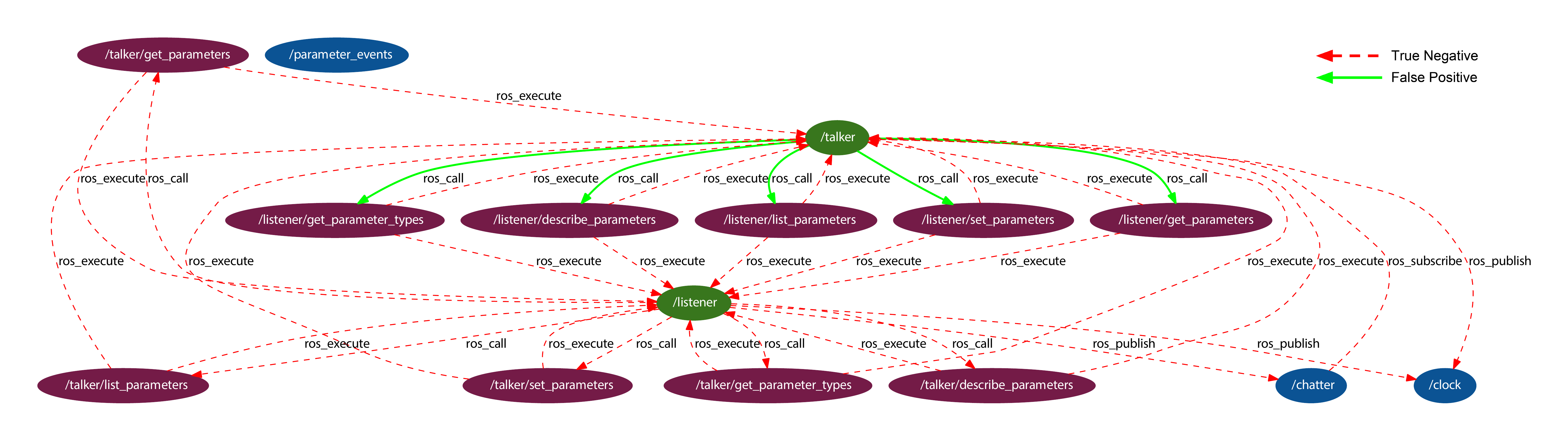}
    \end{subfigure}
    \caption{Annotated graphs depicting transport test results where only the subject $talker$'s policy amended for the empty partition. The colored edges \textcolor{green}{green}/\textcolor{red}{red} correspond to \textcolor{green}{allowed}/\textcolor{red}{denied} actions respectively. Additionally, \textit{True} positive/negative labels are \textit{dashed}, while \textbf{False} positives/negatives are \textbf{solid}. Top graph depicts the intersection between labels from $G_{tp}$ with graph $G_{s}$, while bottom is the relative complement of $G_{s}$ in $G_{tp}$. Given the amendment, $talker$ is now properly capable of connecting to its own objects, as the case is opposed for $listener$ shown via \textbf{solid} edges. However, $talker$ is now also capable of accessing objects intended solely for $listener$. A nuance exists here in that $talker$ and $listener$ share a common service name, though no namespace, any participant with misconfigured QOS/policy exchanging with $listener$ could leak messages to $talker$. Summarizing, any attempted fix that expands the minimal policy set only serves to open new attack surfaces.
    }
    \label{fig:labeled_test_graphs}
\end{sidewaysfigure*}

\newpage
\begin{figure}
    \centering
        \includegraphics[page=3,
            width=\linewidth,
            trim= 0 70 225 35,
            clip]
            {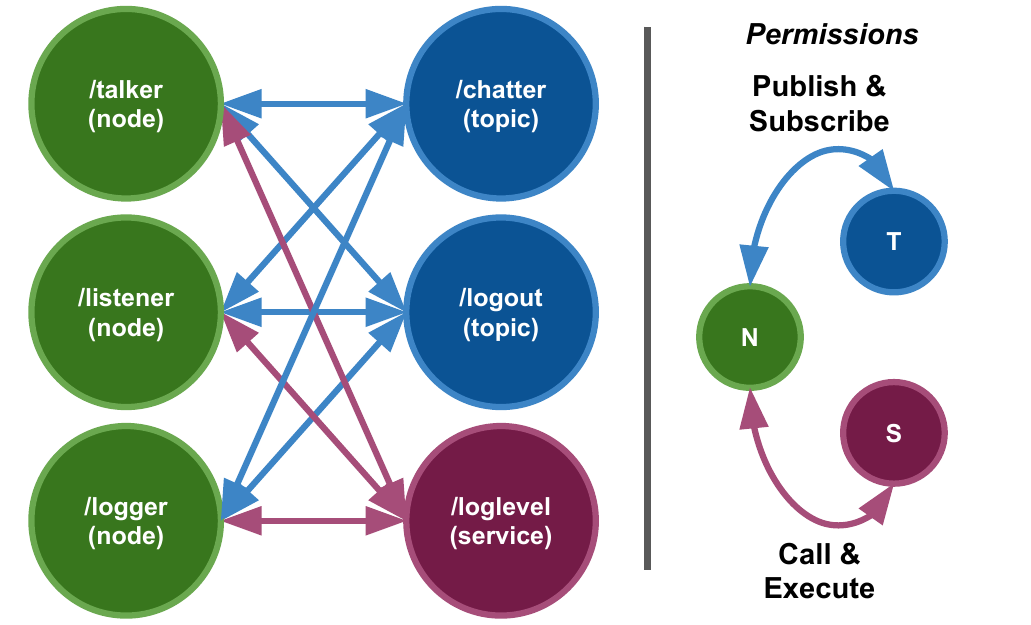}
    \caption{For verification, a complete bipartite graph is formulated from the deployment scenario to provide exhaustive test coverage for each subject and object governed by the policy in question. Vertex pairings are made cyclic as to ensure every permission for each object is also inspected. The graph is then tested both via transport and ComArmor to ensure the allowed and denied edge sets match, and that the allowed set exactly equates to the target deployment scenario.}
    \label{fig:complete_graph}
\end{figure}

\section{Results}
\label{par:results}

For evaluating our framework, a test procedure was developed for verifying the integrity of both the policy and its realized implementation in transport. We begin by generating a semantic graph model $G_{s}$ from an operational yet insecure robotic application, in this case by collecting DDS transport discovery data from the classic ROS2 talker and listener demo publishing and subscribing over the topic chatter. In addition to the topic chatter, many more subsystem level topics are also utilized to extend ROS2 node functionality, the later of which will be shown troubling to properly secure in ROS2 Ardent. 

A minimal satisfactory ComArmor policy $P_{s}$ is then extracted from $G_{s}$. $G_{s}$ is then used again to extrapolate a fully connected bigraph $G_{fc}$, as shown in Figure \ref{fig:complete_graph}, that essentially adds all permissions to all objects for all subjects. We then evaluate graph model $G_{fc}$ semantically using policy $P_{s}$ to classify the edges and output labels $L_{fc}$ for permitted actions.

Next, we generate the transport policies $P_{tp}$ from $P_{s}$ by procedurally compiling with Keymint, in this case manifesting as DDS Security artifacts. $P_{tp}$ is then tested by attempting to deploy $G_{fc}$ using a transport implementation. We then infer the permitted action labels $L_{tp}$ for edges in $G_{fc}$ via logged runtime events from the transport.

Finally, we assert that the set of allowed edges in both $L_{fc}$ and $L_{tp}$ each equate to the set of original edges in $G_{s}$. From the differences between the labels vs original model, the set of false positive (unintended allow) or false negative (unintended deny) for either of the policy tests can be determined. For our case example with the simple talker and listener deployment, experimentation results are shown in Figure \ref{fig:labeled_test_graphs}, where $G_{tp}$ is the graph equivalent of acquired experimental labels $L_{tp}$.

Results show a number of false positives and negative for the transport label set enabling unintended circumvention of the policy by way of cross talk between ROS2 subsystems with namespace omitted. Essentially, this derives from an issue with the current ROS2 Ardent rmw\_connext\_cpp implementation where upon runtime start-up, certain core ROS2 node services are first initialized to the empty string partition. This has resulted in a temporary workaround within SROS2 that simply amends the empty string partition to the list of allowed partition criteria in the transport policy to afford node start up. As of writing, this issue has been ticketed with the ROS2 development team with a resolution forthcoming, anticipated by additional changes in approach to DDS namespace mapping in ROS2 Bouncy release. The source material\footnote{PPAC\_ROS2 Experiments: \href{https://github.com/ruffsl/PPAC\_ROS2}{github.com/ruffsl/PPAC\_ROS2}} for repeating and reproducing our experimental results has also been made available. This experimental material additionally exemplifies a typical work flow using ComArmor and Keymint.

While auditing experimental results, a set of gratuitous permissions within SROS2 default template was also brought to our attention through simple comparative analysis between our minimum spanning policy generated from runtime discovery data and that provided by the SROS2 template. This issue stems from a forgotten holdover workaround in whitelisting DCPS related topics previously necessary for an older DDS security implementation, and has also been ticketed upstream.


\section{Conclusion}
\label{par:conclusion}
In this work we proposed ComArmor as a syntactic access control markup language applicable for robotic systems and computational graphs that leverage channelized hierarchies for invoking actions on subsystem objects. This policy schema was then demonstrated as a proof of concept profile plugin in Keymint, our open source cryptographic meta-build system designed for automating the provisioning of transport specific artifacts enabling secure and controlled communications.

Additionally, we have provided a degree of model verification for semantic permission profiles and a systematic test methodology for checking both application satisfiability and rigour of deployed implementation of a governing policy. The take-away here being the surmountable inherent value of higher level policies that service intermediate representations can provide when coupled with a compile framework to afford transport/vendor agnostic access control definitions.

Although the majority of the model verification has so far assumed finite graphs sizes and policies with exact string attachments, we believe the utilities created during our experimentation will prove helpful for the community, since exhaustive access control evaluation for amenable polices and transport can help bolster unit tests and code coverage.

Lastly, we've covered potential vulnerabilities in SROS2 as encountered over the development and experimentation of our frameworks, emphasizing the importance of continuous security evaluation throughout design development cycle. The authors hope the work presented will be incorporated by the community to help tighten and close the design loop further.

\subsection{Future Work}
\label{par:future_work}

The use of regular expressions within policy profiles offers a powerful and succinct representation to bind subjects and object permissions. Although the default DDS Security access control plugin specification supports a sub set of regular expressions that can be used in binding attachment, and although ComArmor has been designed to support this feature, more formal model verification could be applied to this domain. Even moderate set theory analysis with the restriction to the set of all namespace strings starting or ending with a given sub-string could provide roboticists and system designers greater assurances when using more flexible definitions.     

\begin{figure}
    \centering
        \includegraphics[page=8,
            width=\linewidth,
            trim= 75 45 75 45,
            clip]
            {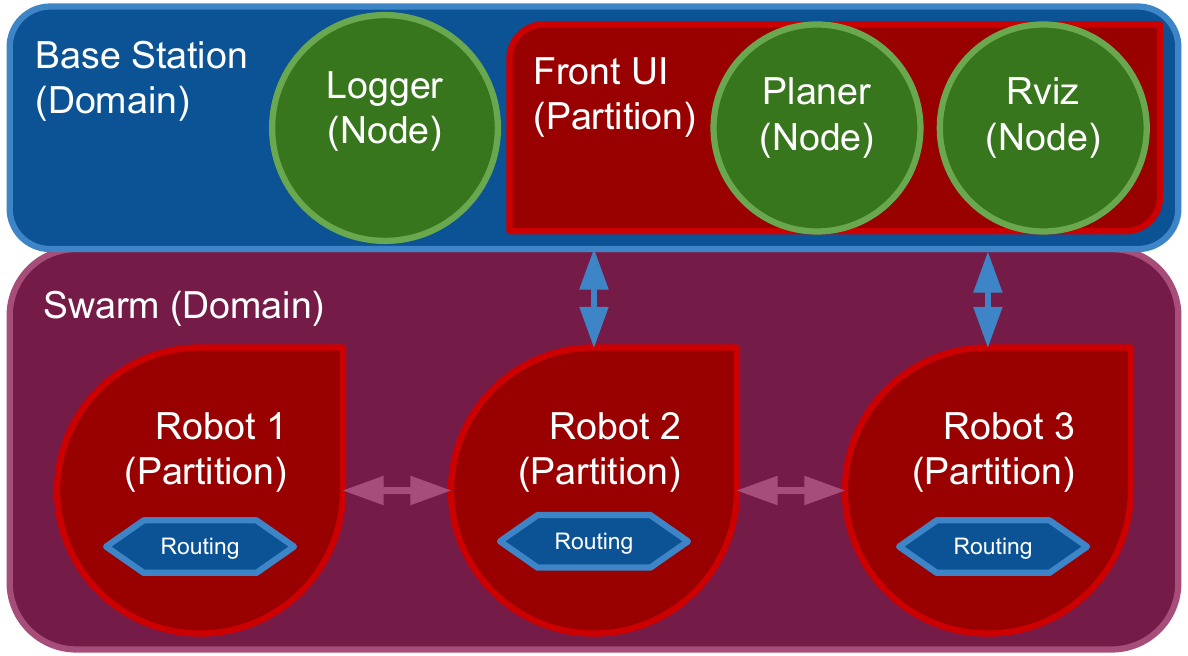}
    \caption{A hypothetical setup for ROS2 enabled swarm leveraging DDS concepts. Traditional ROS namespaces are rooted to an individual robot. Using partitions, robots could switch broadcasts to any or all robots' namespaces. Domains could be used to scope a swarm, utilizing DDS routing for differentiating QOS necessary for intra vs extra swarm communication, such as with lossy remote station vs nearby robot peers.}
    \label{fig:swarm_network}
\end{figure}

Another valuable contribution for secure and scalable robotic deployment would be investigating the design of alternate namespace mapping arrangements in ROS2 that could play to the strengths of DDS, while remaining agnostic to the transport. One such idea would be introducing the concept of $robots$ and $swarms$ into the addressable hierarchy for ROS2 substems (see Figure \ref{fig:swarm_network}) that could serve equally as additional access criteria for broader policy management.


In the authors' opinion, being able to verify the security features at scale is pivotal to quickening and easing progress toward the realization of secure real world robotic products.

\section*{Acknowledgment}

The authors would like to thank the Open Source Robotics Foundation for their support in this work and for the opportunity to contribute to SROS2 and future robot security. Work partially supported by CINI Cybersecurity National Laboratory within the project "FilieraSicura".

\nocite{*}
\bibliographystyle{IEEEtran}
\bibliography{references}

\begin{thebibliography}{10}
\providecommand{\url}[1]{#1}
\csname url@samestyle\endcsname
\providecommand{\newblock}{\relax}
\providecommand{\bibinfo}[2]{#2}
\providecommand{\BIBentrySTDinterwordspacing}{\spaceskip=0pt\relax}
\providecommand{\BIBentryALTinterwordstretchfactor}{4}
\providecommand{\BIBentryALTinterwordspacing}{\spaceskip=\fontdimen2\font plus
\BIBentryALTinterwordstretchfactor\fontdimen3\font minus
  \fontdimen4\font\relax}
\providecommand{\BIBforeignlanguage}[2]{{%
\expandafter\ifx\csname l@#1\endcsname\relax
\typeout{** WARNING: IEEEtran.bst: No hyphenation pattern has been}%
\typeout{** loaded for the language `#1'. Using the pattern for}%
\typeout{** the default language instead.}%
\else
\language=\csname l@#1\endcsname
\fi
#2}}
\providecommand{\BIBdecl}{\relax}
\BIBdecl

\bibitem{Mineraud2016}
\BIBentryALTinterwordspacing
J.~Mineraud, O.~Mazhelis, X.~Su, and S.~Tarkoma, ``{A gap analysis of
  Internet-of-Things platforms},'' \emph{Computer Communications}, vol. 89-90,
  pp. 5--16, 2016. [Online]. Available:
  \url{http://dx.doi.org/10.1016/j.comcom.2016.03.015}
\BIBentrySTDinterwordspacing

\bibitem{Morante2015}
\BIBentryALTinterwordspacing
S.~Morante, J.~G. Victores, and C.~Balaguer, ``{Cryptobotics: Why Robots Need
  Cyber Safety},'' \emph{Frontiers in Robotics and AI}, vol.~2, no. September,
  pp. 23--26, sep 2015. [Online]. Available:
  \url{http://journal.frontiersin.org/Article/10.3389/frobt.2015.00023/abstract}
\BIBentrySTDinterwordspacing

\bibitem{Cardenas}
A.~A. C{\'{a}}rdenas, S.~Amin, B.~Sinopoli, A.~Giani, A.~Perrig, and S.~Sastry,
  ``{Challenges for Securing Cyber Physical Systems}.''

\bibitem{Dieber2017}
\BIBentryALTinterwordspacing
B.~Dieber, B.~Breiling, S.~Taurer, S.~Kacianka, S.~Rass, and P.~Schartner,
  ``{Security for the Robot Operating System},'' \emph{Robotics and Autonomous
  Systems}, vol.~98, pp. 192--203, dec 2017. [Online]. Available:
  \url{https://doi.org/10.1016/j.robot.2017.09.017
  http://linkinghub.elsevier.com/retrieve/pii/S0921889017302762}
\BIBentrySTDinterwordspacing

\bibitem{quigley2009ros}
M.~Quigley, K.~Conley, B.~Gerkey, J.~Faust, T.~Foote, J.~Leibs, R.~Wheeler, and
  A.~Y. Ng, ``{ROS}: an open-source robot operating system,'' in \emph{ICRA
  workshop on open source software}, vol.~3, no. 3.2.\hskip 1em plus 0.5em
  minus 0.4em\relax Kobe, Japan, 2009, p.~5.

\bibitem{pardo2003}
G.~Pardo-Castellote, ``Omg data-distribution service: architectural overview,''
  in \emph{23rd International Conference on Distributed Computing Systems
  Workshops, 2003. Proceedings.}, May 2003, pp. 200--206.

\bibitem{Denning2009}
T.~Denning, C.~Matuszek, K.~Koscher, J.~R. Smith, and T.~Kohno, ``A spotlight
  on security and privacy risks with future household robots: Attacks and
  lessons,'' in \emph{Proceedings of the 11th International Conference on
  Ubiquitous Computing}, ser. UbiComp '09, 2009, pp. 105--114.

\bibitem{Lee2012}
\BIBentryALTinterwordspacing
G.~S. Lee and B.~Thuraisingham, ``{Cyberphysical systems security applied to
  telesurgical robotics},'' \emph{Computer Standards {\&} Interfaces}, vol.~34,
  no.~1, pp. 225--229, jan 2012. [Online]. Available:
  \url{http://dx.doi.org/10.1016/j.csi.2011.09.001
  http://linkinghub.elsevier.com/retrieve/pii/S0920548911000870}
\BIBentrySTDinterwordspacing

\bibitem{huang2014}
J.~Huang, C.~Erdogan, Y.~Zhang, B.~Moore, Q.~Luo, A.~Sundaresan, and G.~Rosu,
  ``Rosrv: Runtime verification for robots,'' in \emph{Proceedings of the 14th
  International Conference on Runtime Verification}, ser. LNCS, vol.
  8734.\hskip 1em plus 0.5em minus 0.4em\relax Springer International
  Publishing, September 2014, pp. 247--254.

\bibitem{Doczi2016}
R.~Dóczi, F.~Kis, B.~Sütő, V.~Póser, G.~Kronreif, E.~Jósvai, and
  M.~Kozlovszky, ``Increasing ros 1.x communication security for medical
  surgery robot,'' in \emph{2016 IEEE International Conference on Systems, Man,
  and Cybernetics (SMC)}, Oct 2016, pp. 4444--4449.

\bibitem{portugal2017}
D.~Portugal, M.~A. Santos, S.~Pereira, and M.~S. Couceiro, ``On the security of
  robotic applications using {ROS},'' in \emph{Artificial Intelligence Safety
  and Security}.\hskip 1em plus 0.5em minus 0.4em\relax CRC Press, December
  2017.

\bibitem{white2016sros}
R.~White, M.~Quigley, and H.~Christensen, ``{SROS}: Securing {ROS} over the
  wire, in the graph, and through the kernel,'' in \emph{Humanoids Workshop:
  Towards Humanoid Robots {OS}}.\hskip 1em plus 0.5em minus 0.4em\relax Cancun,
  Mexico, 2016.

\bibitem{white16roscon}
\BIBentryALTinterwordspacing
R.~White and M.~Quigley, ``{\{,S\}ROS}: Securing {ROS} over the wire, in the
  graph, and through the kernel,'' 2016, {ROSCon}, Seoul South Korea. [Online].
  Available: \url{https://vimeo.com/187705073}
\BIBentrySTDinterwordspacing

\bibitem{SecureROS}
\BIBentryALTinterwordspacing
M.~K. A.~Sundaresan, L.~Gerard. (2017) Secure ros 0.9.2 documentation.
  [Online]. Available: \url{https://sri-csl.github.io/secure_ros}
\BIBentrySTDinterwordspacing

\bibitem{dieber2016application}
B.~Dieber, S.~Kacianka, S.~Rass, and P.~Schartner, ``Application-level security
  for {ROS}-based applications,'' in \emph{Intelligent Robots and Systems
  (IROS), 2016 IEEE/RSJ International Conference on}.\hskip 1em plus 0.5em
  minus 0.4em\relax IEEE, 2016, pp. 4477--4482.

\bibitem{breiling2017secure}
B.~Breiling, B.~Dieber, and P.~Schartner, ``Secure communication for the robot
  operating system,'' in \emph{2017 Annual IEEE International Systems
  Conference (SysCon)}, April 2017, pp. 1--6.

\bibitem{Koubaa2018}
R.~White, G.~Caiazza, H.~Christensen, and A.~Cortesi, ``{SROS1}: Using and
  developing secure {ROS1} system,'' in \emph{Robot Operating System (ROS): The
  Complete Reference (Volume 3)}.\hskip 1em plus 0.5em minus 0.4em\relax
  Springer, to appear, 2018.

\bibitem{McClean2013}
\BIBentryALTinterwordspacing
J.~McClean, C.~Stull, C.~Farrar, and D.~Mascare{\~{n}}as, ``{A preliminary
  cyber-physical security assessment of the Robot Operating System (ROS)},''
  vol. 8741, p. 874110, 2013. [Online]. Available:
  \url{http://proceedings.spiedigitallibrary.org/proceeding.aspx?doi=10.1117/12.2016189}
\BIBentrySTDinterwordspacing

\bibitem{CortesiFC13}
A.~Cortesi, P.~Ferrara, and N.~Chaki, ``Static analysis techniques for robotics
  software verification,'' in \emph{Proceedings of the 44th Internationel
  Symposium on Robotics, {IEEE} {ISR} 2013, Seoul, Korea (South), October
  24-26, 2013}, 2013, pp. 1--6.

\bibitem{bauer2006paranoid}
M.~Bauer, ``Paranoid penguin: an introduction to novell apparmor,'' \emph{Linux
  Journal}, vol. 2006, no. 148, p.~13, 2006.

\bibitem{hohenberger2014online}
S.~Hohenberger and B.~Waters, ``Online/offline attribute-based encryption,'' in
  \emph{International Workshop on Public Key Cryptography}.\hskip 1em plus
  0.5em minus 0.4em\relax Springer, 2014, pp. 293--310.

\bibitem{lera2016}
F.~J.~R. Lera, J.~Balsa, F.~Casado, C.~Fern{\'a}ndez, F.~M. Rico, and
  V.~Matell{\'a}n, ``Cybersecurity in autonomous systems: Evaluating the
  performance of hardening {ROS},'' \emph{M{\'a}laga, Spain-June 2016}, p.~47,
  2016.

\end{thebibliography}

\end{document}